\documentclass[letterpaper]{article} 
\usepackage{aaai24}  
\usepackage{times}  
\usepackage{helvet}  
\usepackage{courier}  
\usepackage[hyphens]{url}  
\usepackage{graphicx} 
\usepackage{natbib}  
\usepackage{caption} 
\usepackage{algorithm}
\usepackage{algorithmic}

\usepackage{newfloat}
\usepackage{listings}
\DeclareCaptionStyle{ruled}{labelfont=normalfont,labelsep=colon,strut=off} 
\floatstyle{ruled}
\newfloat{listing}{tb}{lst}{}
\floatname{listing}{Listing}
\usepackage{amsmath,amsthm,amsfonts,bm,amssymb}
\usepackage[parfill,skip=15pt]{parskip}
\usepackage{thm-restate}

\newtheorem{kernel}{Kernel}

\DeclareMathOperator*{\argmin}{\arg\!\min}
\newcommand\numberthis{\addtocounter{equation}{1}\tag{\theequation}}

\DeclareMathOperator\ident{Id}
\DeclareMathOperator{\Tr}{Tr}

\renewcommand{\vector}[1]{\bm{\lowercase{#1}}}
\renewcommand{\matrix}[1]{\bm{\uppercase{#1}}}
\newcommand{\rv}[1]{\mathsf{#1}}

\usepackage[table,x11names]{xcolor}
\usepackage{collcell}
\usepackage{array}
\usepackage{tikz}
\newcolumntype{H}{>{\collectcell\Heat}r<{\endcollectcell}}
\usepackage{multirow}

\usepackage{pifont}

\newcommand{\russ}[1]{\textcolor{blue}{[{\sc russ}] #1}}

\title{Exact, Fast and Expressive Poisson Point Processes via Squared Neural Families}
\author{
    Russell Tsuchida\textsuperscript{\rm 1},
    Cheng Soon Ong\textsuperscript{\rm 1, \rm 2},
    Dino Sejdinovic\textsuperscript{\rm 3}
}
\affiliations{
    \textsuperscript{\rm 1}Data61-CSIRO \\
    \textsuperscript{\rm 2}Australian National University \\
    \textsuperscript{\rm 3}University of Adelaide
}

\begin{document}

\maketitle

\begin{abstract}
We introduce squared neural Poisson point processes (SNEPPPs) by parameterising the intensity function by the squared norm of a two layer neural network.
When the hidden layer is fixed and the second layer has a single neuron, our approach resembles previous uses of squared Gaussian process or kernel methods, but allowing the hidden layer to be learnt allows for additional flexibility.
In many cases of interest, the integrated intensity function admits a closed form and can be computed in quadratic time in the number of hidden neurons.
We enumerate a far more extensive number of such cases than has previously been discussed.
Our approach is more memory and time efficient than naive implementations of squared or exponentiated kernel methods or Gaussian processes.
Maximum likelihood and maximum a posteriori estimates in a reparameterisation of the final layer of the intensity function can be obtained by solving a (strongly) convex optimisation problem using projected gradient descent. 
We demonstrate SNEPPPs on real, and synthetic benchmarks, and provide a software implementation.
\end{abstract}
\section{Introduction}

\paragraph{Intensity Functions}
Our goal is to develop machine learning methods for learning intensity functions, which are central building blocks for probabilistic models of points in sets.
An intensity function $\lambda$ quantifies the number of points $\rv{N}(A)$ in a set $A \subseteq \mathbb{X}$ divided by the size of the set, as the size of the set approaches zero. 
In other words, $\lambda$ is a nonnegative function that may be integrated over the set $A$ against some measure $\mu$ to give the expected number of points.
We give formal definition of intensity functions and inhomogeneous Poisson point processes (PPPs) in \S~\ref{sec:ppps}.
PPPs and their extensions find a wide range of applications in, e.g., seismology~\citep{ogata1988statistical}, ecology~\citep{renner2015point} and popularity prediction in social networks~\citep{mishra2016feature}.
The domain $\mathbb{X}$ may be of arbitrary dimension $d$, although much attention in statistics has been given to the one, two, three and four dimensional settings, which are often identified with the qualifiers ``queueing'' or ``temporal'' (1D), ``spatial'' (2D or 3D) or ``spatio-temporal'' (3D or 4D). 
Here we consider the case of arbitrary dimension $d$.

\begin{figure}[t]
    \centering
    \includegraphics[scale=0.05]{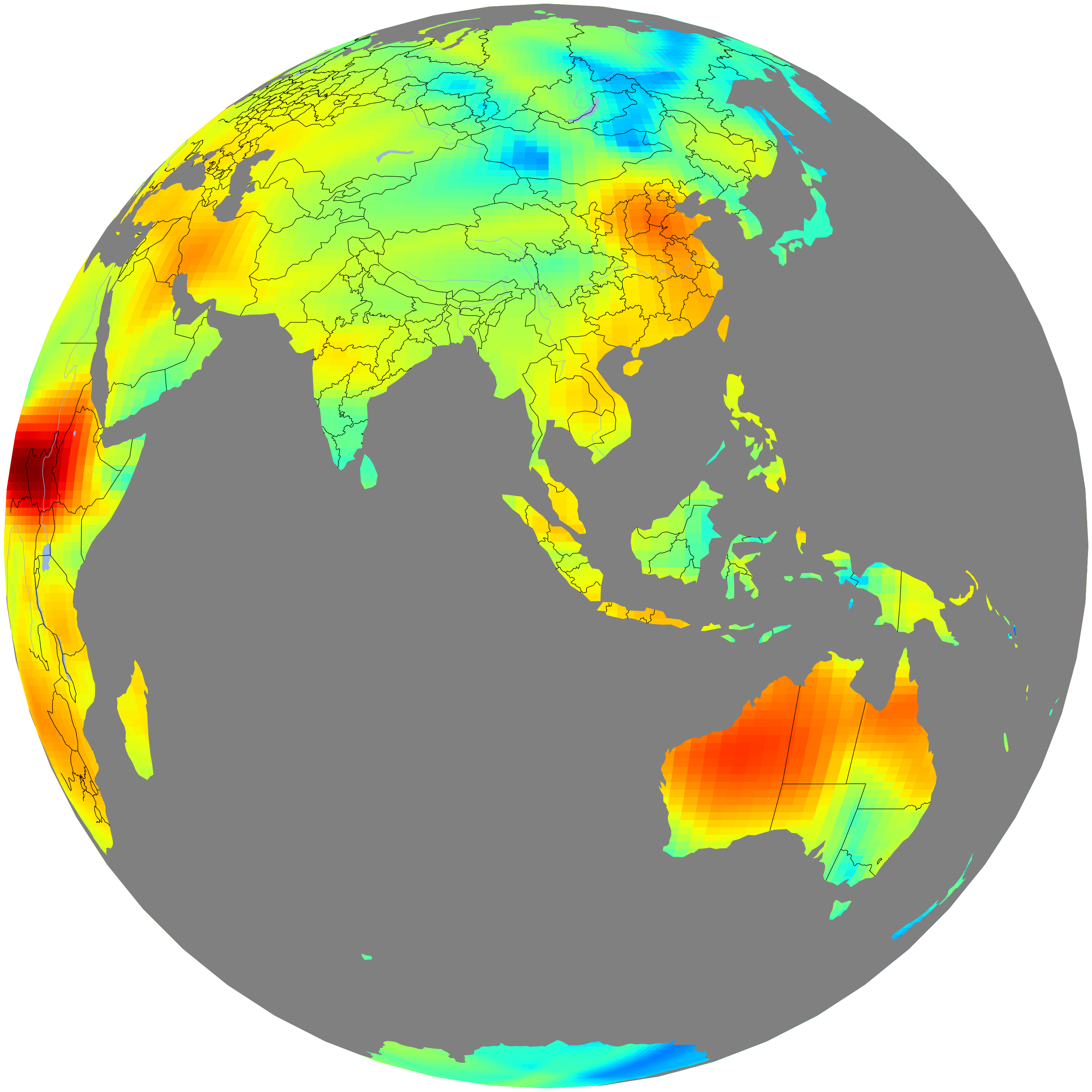}
    \includegraphics[scale=0.05]{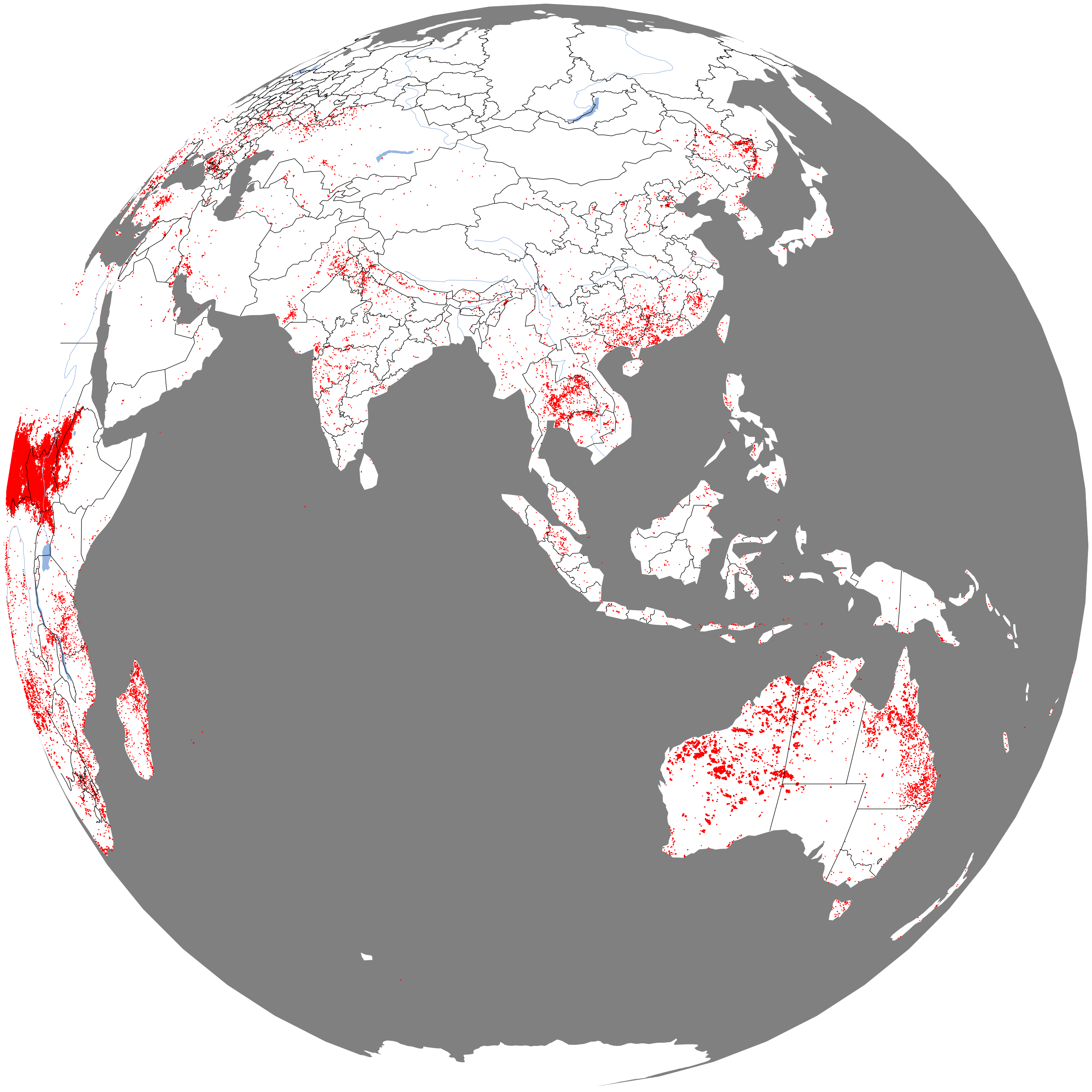}
    \caption{A SNEPPP spatial intensity function fit (left) to 200,000 events using NASA wildfire data (right)~\citep{bushfires}. In \S~\ref{sec:wildfires}, we discuss a video of a spatio-temporal fit using 100 million events.}
    \label{fig:wildfires_globe}
\end{figure}

\paragraph{Desiderata for Intensity Functions} 
We discuss three useful properties that intensity functions should possess: \emph{expressivity}, \emph{tractable integration} and \emph{tractable optimisation}.
Firstly, one needs to choose a function class $\mathcal{H}$ from which to represent the intensity function $\lambda \in \mathcal{H}$.
This function class should be \emph{expressive} enough to represent the data at hand.
Importantly, $\lambda$ must be nonnegative, which precludes direct application of some of the standard choices of $\mathcal{H}$ such as RKHSs or certain neural networks.
Secondly, computing the expected number of points in a set requires integrating the intensity function, as does measuring the likelihood of a given realisation from a PPP.
The integrated intensity function therefore needs to be evaluated often in a typical model fitting and testing pipeline.
\emph{Computing or approximating the integrated intensity function} can be challenging for all but simple models and observation windows.
Finally, even if the integrated intensity function can be computed, finding plausible members of the function class $\mathcal{H}$ can be difficult.
Even point estimation (i.e. choosing a single ``best'' $\lambda \in \mathcal{H}$) according to some optimisation criterion can be hard if the \emph{optimisation criterion is nonconvex}, which is typically the case for non-trivial hypothesis classes $\mathcal{H}$.

\paragraph{Contributions}
We introduce a new class of intensity functions that addresses each of the three desiderata mentioned above.
We use the squared Euclidean norm of a two layer neural network to parameterise the intensity function.
This represents a highly flexible universal approximating function class, with even more flexibility if the hidden layer is also trained.
Under this function class, the integrated intensity function can be computed exactly for a very wide range of neural network activation functions. 
In order to compute the integrated intensity function, we extend a recently developed framework for modelling probability densities~\citep{tsuchida2023squared}.
Unlike naive kernel methods which scale cubically in the number of datapoints in computing the integrated intensity function, our method scales only in the dimensionality of the feature mappings.
Finally, we show that if the hidden layer is not trained, the classical maximum likelihood estimate (MLE) and maximum a posteriori (MAP) estimate can be found by minimising a (strongly) convex function over a convex set. 
Such estimates can be found using projected gradient descent.
The performance and features of our method are illustrated on some real and synthetic benchmark datasets.
We empirically demonstrate the efficacy and efficiency of our open source method on a large scale case study of wildfire data from NASA with about 100 million events.

\section{Background}

\subsection{Poisson Point Processes}
\label{sec:ppps}
We briefly present definitions for PPP likelihoods and intensity functions, referring the reader to classical sources for more details~\citep[\S~8.5]{Cressie1993}~\citep[\S~4.3]{Baddeley2007}.
Let $\mathbb{X} \subseteq \mathbb{R}^d$ with Borel sigma algebra $\mathcal{F}$.
Let $\nu$ be a nonnegative measure such that $\nu(A)<\infty$ for any compact $A \in \mathcal{F}$.
We call $\nu$ the intensity measure.
An (inhomogeneous) \emph{Poisson point process} (PPP) $\rv{N}$ on $\mathbb{X}$ specifies that for any such $A$, the number $\rv{N}(A)$ of events within $A$ is governed by a Poisson distribution with a parameter equal to the evaluated intensity measure $\nu(A)>0$,
\begin{align*}
    \mathbb{P}\big( \rv{N}(A) = N\big) = \frac{\big(\nu(A) \big)^N\exp(-\nu(A))}{N!}. \label{eq:poisson} \numberthis 
\end{align*}
Let $\vector{\rv{x}}_i$ denote one of the $\rv{N}(A)$ points in $A$.
Conditional on the event $\rv{N}(A) = N$, the ordered $N$-tuple $(\vector{\rv{x}}_1,\ldots,\vector{\rv{x}}_N)$, is distributed independently and identically with probability distribution $\nu(\cdot) / \nu(A)$. If $\nu$ is absolutely continuous with respect to some \emph{base measure} $\mu$ and $\lambda=\frac{d\nu}{d\mu}:\mathbb{X} \to [0, \infty)$ is the Radon-Nikodym derivative called the \emph{intensity function}, the random vector $(\vector{\rv{x}}_1,\ldots,\vector{\rv{x}}_N)$ admits a conditional density function with respect to $\mu$
\begin{align*}
p(\vector{x}_1,\ldots,\vector{x}_N|\rv{N}(A)=N,\lambda)=\frac{\prod_{i=1}^N\lambda(\vector{x}_i)}{\big(\int_{A}\lambda(\vector{u}) \mu(d\vector{u}) \big)^N}. \label{eq:conditional_points} \numberthis
\end{align*}
Finally, combining~\eqref{eq:poisson} and~\eqref{eq:conditional_points}, the \emph{joint likelihood} of the events and the number $\rv{N}(A)$ of the events is
\begin{align*}
&\phantom{{}={}}p(\vector{x}_1,\ldots,\vector{x}_N, N|\lambda)=\\
&\frac{1}{N!}\prod_{i=1}^{N}\lambda(\vector{x}_i)\exp\left(-\int_{A}\lambda(\vector{u}) \mu(d\vector{u}) \right). \numberthis \label{eq:joint}
\end{align*}

\subsection{Squared Neural Families}

One way to construct a probability density function (with respect to some base measure) is to divide a nonnegative function by its integral (with respect to a base measure). 
The challenge in this method is computing the integral, also called the normalising constant. 
\citet{tsuchida2023squared} construct families of probability distributions by normalising the squared Euclidean norm of two layer neural networks. 
They show that such a procedure often results in a closed-form normalising constant in terms of a neural network kernel (NNK).
We will use an analogous result in the context of Poisson point processes instead of probability distributions, see Identity~\ref{identity:trace}.

\section{Flexible Intensity with a Tractable Integral}
\subsection{Squared Neural Poisson Point Process}
\paragraph{Our Parameterisation of Intensity Function $\lambda$}
We investigate the use of finite features for the intensity.
We begin by modelling the intensity function as the product of a hyperparameter $0< \alpha <\infty$ and the squared Euclidean norm of a finite feature model with \emph{fixed} features $\vector{\psi}$,
\begin{subequations}\label{eq:intensity}
\begin{align*}
    \lambda(\vector{x}) = \alpha\big\Vert \matrix{V} \vector{\psi}(\vector{x})\big\Vert_2^2, \numberthis \label{eq:intensity1}
\end{align*}
where $\matrix{V} \in \mathbb{R}^{m \times n}$ and $\vector{\psi}:\mathbb{R}^d \to \mathbb{R}^n$.
Examples of such $\vector{\psi}$ include radial basis function networks with fixed hidden layer parameters~\citep[which are universal approximators]{broomhead1988radial}, random Fourier features~\citep[which can approximate universal kernels]{rahimi2007random} and finite dimensional kernel feature mappings~\citep[definition 9.1, 9.5, 9.6 and others]{shawe2004kernel}.
Shallow and deep neural networks with fixed (but possibly randomly sampled) hidden parameters also fall under~\eqref{eq:intensity1}~\citep[for example, (8) and (9)]{NIPS2009_3628}.
Fixed features are previously used in the context of PPPs when $m=1$ by~\citet{walder2017fast}, however our extension to $m\geq 1$ not only increases flexibility but also allows us to prove optimisation properties (see Proposition~\ref{prop:optimisation}).

We then set the features $\vector{\psi}$ to be a hidden neural network layer with learnable hidden parameters.
Suppose the hidden activation function is $\sigma$ (applied element-wise), and input data $\vector{x} \in \mathbb{R}^d$ is preprocessed by passing through a warping function $\vector{t}:\mathbb{R}^d \to \mathbb{R}^D$. 
We set the intensity function $\lambda$ to be equal to the squared Euclidean norm of the neural network's output multiplied by a hyperparameter $0< \alpha < \infty$,
\begin{align*}
    \lambda(\vector{x}) = \alpha\big\Vert \matrix{V} \vector{\psi}(\vector{x})\big\Vert_2^2, \quad \vector{\psi}(\vector{x}) = \sigma(\matrix{w} \vector{t}(\vector{x}) + \vector{B})), \numberthis \label{eq:intensity2}
\end{align*}
\end{subequations}
where we introduced additional parameters to be learned: hidden layer weights and biases $\matrix{W} \in \mathbb{R}^{n \times D}$ and $\vector{b} \in \mathbb{R}^n$.
For fixed $(\matrix{W}, \vector{b})$,~\eqref{eq:intensity2} is a special case of~\eqref{eq:intensity1}.

\paragraph{Extension to Product Spaces} We provide one additional extension of~\eqref{eq:intensity1} and~\eqref{eq:intensity2} in Appendix~B, 
which is particularly useful when the space $\mathbb{X}$ naturally decomposes into a Cartesian product (for example, as one might be interested in when studying spatio-temporal models).
For the purposes of notational simplicity, we discuss our central ideas in terms of models~\eqref{eq:intensity1} and~\eqref{eq:intensity2}, noting that these ideas naturally extend to the model Appendix~B. 
Our large scale case study in \S~\ref{sec:experiments} uses the setting in Appendix~B.

\paragraph{Integrated Intensity Function}
One challenge is in computing the integrated intensity function,
$$\Lambda= \int_{A} \lambda(\vector{x}) \mu(d\vector{x}) = \alpha \int_{A} \big\Vert \matrix{V} \vector{\psi}(\vector{x}) \big\Vert_2^2 \mu(d\vector{x}) , \qquad A \subseteq \mathbb{X},$$
focusing mostly on the setting where $A = \mathbb{X}$ (i.e. the process is observed over a full window)\footnote{We are also able to handle cases where $A$ is ``full'' in a subset of dimensions. For example, in the spatio-temporal setting described in \S~\ref{sec:spatio-temporal}, we integrate over a full spatial domain and a partially observed temporal domain.}. 
We note that a setting amounting to full observation window is also considered by other machine learning approaches~\citep{flaxman2017poisson,walder2017fast}.
Much of the machinery we develop here mirrors the use of squared neural families for density estimation~\citep{tsuchida2023squared}, however we introduce a number of new kernels and results specific to the setting of intensity estimation and aim to give a self-contained presentation.

\paragraph{Regularisation} Note that either~\eqref{eq:intensity1} or~\eqref{eq:intensity2} implies that the conditional likelihood~\eqref{eq:conditional_points} does not depend on $\alpha$, but the joint likelihood~\eqref{eq:joint} does depend on $\alpha$.
The inclusion of $\alpha$, borrowed from~\citet[see remark 2]{flaxman2017poisson}, allows us to explicitly control the overall scale of the intensity independently from other measures of complexity of the intensity (such as the parameter Euclidean norm, as might typically be done using $\ell^2$ regularisation).
This can be helpful for model fitting. 
Intuitively speaking, if $\Vert \matrix{V} \vector{\psi}(\vector{x}) \Vert^2$ is initialised to have the value $1$, and we observe $N$ samples over a volume of $\mu(A)$, $\alpha$ should be set to roughly $N/\mu(A)$ to encourage the initialised model to predict the correct number of training observations. 

\paragraph{Bayesian Extensions}
One may place a Bayesian prior over parameters of $\lambda$ and compute with the resulting posterior. 
For example, if one were to employ a Gaussian prior over $\matrix{V}$, then $\matrix{V} \vector{\psi}(\cdot)$ would be a vector-valued Gaussian process prior, extending previously considered models~\citep{walder2017fast}.
While we do not consider such Bayesian models in detail here, we mention related work in \S~\ref{sec:related}.
We compute MAP estimates in Proposition~\ref{prop:optimisation}.


\subsection{Closed Form Integration}
\label{sec:iif}
The following identity is proven in Appendix~A. 
It is a variation of the result in~\citet{tsuchida2023squared}, the difference being that here $\lambda$ is an intensity function rather than an unnormalised probability density function.
\begin{restatable}{proposition}{trace}\label{identity:trace} 
Under~\eqref{eq:intensity1} (and therefore~\eqref{eq:intensity2}), the intensity function and integrated intensity function are
\begin{align*}
    \lambda(\vector{x}) = \alpha \Tr \Big(  \matrix{V}^\top \matrix{V} \widetilde{\matrix{K}}(\vector{x})  \Big) \quad \text{and} \quad \Lambda = \alpha \Tr\Big(\matrix{V}^\top \matrix{V} \matrix{K}\Big),
\end{align*}
where $\widetilde{\matrix{K}}(\vector{x})$ is the PSD matrix with $ij$th entry $\psi_i(\vector{x})\psi_j(\vector{x})$ and $\matrix{K}$ is the PSD matrix with $ij$th entry $\kappa(i,j)$,
\begin{align*}
    \kappa(i,j) = \int_{\mathbb{X}} \psi_i(\vector{x}) \psi_j(\vector{x}) \mu(d\vector{x}).
\end{align*}
\end{restatable}

\paragraph{Neural Network Kernels}
In the special case of~\eqref{eq:intensity2}, we identify $\kappa$ with a broad family of kernels called neural network kernels (NNKs).
We let $\matrix{\Theta} = (\matrix{W}, \vector{B}) \in \mathbb{R}^{n\times (D+1)}$, and denote the $i$th row of $\matrix{\Theta}$ by $\vector{\theta}_i = (\vector{w}_i, b_i) \in \mathbb{R}^{D+1}$.
In this case, it is more natural to consider the kernel as operating over the space of parameters $\vector{\theta}_i, \vector{\theta}_j$ indexed by $i, j$.
We define the NNK $k_{\sigma,\vector{t},\mu}(\vector{\theta}_i, \vector{\theta}_j)$ to be such a kernel,
\begin{align*}
\kappa(i,j) &= \int_{\mathbb{X}} \sigma\big(\vector{W}_i^\top \vector{t}(\vector{x}) + b_i\big) \sigma\big(\vector{W}_j^\top \vector{t}(\vector{x}) + b_j\big) \mu(d\vector{x}) \\
&=: k_{\sigma,\vector{t},\mu}(\vector{\theta}_i, \vector{\theta}_j). \label{eq:general_nnk} \numberthis
\end{align*}

\paragraph{Closed Forms and Computational Complexity} If we have a closed form for the kernel $\kappa$, we may compute $\Lambda=\Lambda(\matrix{V}, \matrix{\Theta})$ in $\mathcal{O}(m^2n + n^2)$.
Closed forms for the NNK $k_{\sigma,\vector{t},\mu}(\vector{\theta}_i, \vector{\theta}_j)$ for various choices of $\sigma, \vector{t}$ and $\mu$ are available; see ~\citet[table 1]{han2022fast} and~\citet[table 1]{tsuchida2023squared} and references therein. 
These NNKs allow the use of a variety of activation functions, including error functions, ReLU, Leaky ReLU, GELU~\citep{hendrycks2016gaussian}, snake~\citep{NEURIPS2020_11604531}, Gaussians, polynomials, and Gabor functions.
Each evaluation of such NNK typically incurs a cost of $\mathcal{O}(d)$, depending on $\vector{w}_j$ and $\vector{w}_j$ only through a Euclidean inner product between $\vector{\theta}_i$ and $\vector{\theta}_j$.
This linear complexity in input dimension $d$ is extremely favourable in light of the exponential complexity that would be implied by a naive numerical integration technique.
It is also favourable compared with kernel methods that use the representer theorem, which scale cubically in the number of datapoints $N$.

\subsection{Novel Examples of NNKs}
\label{sec:nnk_examples}
In \S~\ref{sec:iif}, we reduced calculation of the integrated intensity function to calculation of the NNK $k_{\sigma,\vector{t},\mu}(\vector{\theta}_i, \vector{\theta}_j)$, and pointed to a vast literature that provides closed-forms for NNKs.
Below we discuss two other examples of NNKs not mentioned in the aforementioned works.

\paragraph{Exponential Family Type} Take an identity warping function $\vector{t}(\vector{x}) = \vector{x} = \ident(\vector{x})$, a hyperrectangular domain $\mathbb{X} = [a,b]^d$, an exponential activation function $\sigma = \exp$ and the base measure to be Lebesgue on $[a,b]^d$.
The NNK can then be computed in closed-form,
\begin{align*}
    k_{\exp,\ident,d\vector{x}}(\vector{\theta}_i, \vector{\theta}_j) &= e^{b_i + b_j} \int_{[a,b]^d} \exp\big(\vector{x}^\top (\vector{w}_i + \vector{w}_j) \big) d\vector{x} \\
    &= e^{b_i + b_j} \prod_{r=1}^d \Bigg(\frac{\exp\big( x_r (w_{ir} + w_{jr}) \big)}{w_{ir} + w_{jr}} \Big|^b_a \Bigg).
\end{align*}
In this case, the NNK $k_{\exp,\ident,d\vector{x}}(\vector{\theta}_i, \vector{\theta}_j)$ also defines a partition function of an exponential family~\citep{wainwright2008graphical,nielsen2009statistical} with identity sufficient statistic supported on $[a,b]^d$.
In a similar vein, other exponential families can define analogous SNEPPPs.

\paragraph{Absolutely Homogeneous Activations on the Sphere} 
A function $\sigma^{(p)}$ is said to be absolutely $p$-homogeneous if for some $p >0$, for all $a$ and $z$, $\sigma^{(p)}(|a|z) = |a|^p \sigma^{(p)}(z)$.
Examples of absolutely $p$-homogeneous functions include monomials, the Heaviside step function, the ReLU (raised to any positive power), and the Leaky ReLU.

If the bias parameter $\vector{b} = \vector{0}$, the activation $\sigma=\sigma^{(p)}$ is $p$-homogeneous, and the warping function $\vector{t}(\vector{x})=\vector{x}$ is the identity, we may translate between NNKs with uniform base measure on the sphere and NNKs with Gaussian base measure on $\mathbb{R}^d$, as follows. 
It is well-known that if $\vector{\rv{x}}$ is zero mean isotropic Gaussian, then $\vector{\rv{x}}/\Vert \vector{\rv{x}} \Vert$ and $\Vert \vector{\rv{x}} \Vert$ are independent. Using this fact and the absolute homogeneity property,
\begin{align*}
    k_{\sigma^{(p)}, \ident,\Phi}(\vector{\theta}_i, \vector{\theta}_j) &= \mathbb{E}[\sigma^{(p)}(\vector{w}_i^\top \vector{\vector{\rv{x}}})\sigma^{(p)}(\vector{w}_j^\top \vector{\vector{\rv{x}}})] \\
    &= \mathbb{E}\big[ \Vert \vector{\vector{\rv{x}}}\Vert^{2p} \sigma^{(p)}\big(\vector{w}_i^\top \frac{\vector{\vector{\rv{x}}}}{\Vert \vector{\vector{\rv{x}}} \Vert} \big)\sigma^{(p)}\big(\vector{w}_j^\top \frac{\vector{\vector{\rv{x}}}}{\Vert \vector{\vector{\rv{x}}} \Vert}\big)\big] \\
    &=  \mathbb{E}[ \Vert \vector{\vector{\rv{x}}}\Vert^{2p} ] k_{\sigma^{(p)}, \ident, U(\mathbb{S}^{d-1})}(\vector{\theta}_i, \vector{\theta}_j). \numberthis \label{eq:gauss2sphere}
\end{align*}
Finally, $\mathbb{E}[ \Vert \vector{\vector{\rv{x}}}\Vert^{2p} ]$ is just the $p$th moment of a Chi-squared random variable, which is available in closed-form.

As an example, consider the family of activation functions of the form $\sigma(z) = \Theta(z) z^p$, where $\Theta$ is the Heaviside step function. 
The ReLU is obtained when $p=1$. 
The arc-cosine kernel of order $p$ is known in closed form if $\vector{b}=\vector{0}$~\citep{NIPS2009_3628}.
In particular, when $p=1$,
\begin{align*}
    k_{(\cdot)\Theta(\cdot) , \ident,\Phi}(\vector{\theta}_i, \vector{\theta}_j) = \frac{\Vert \vector{w}_i \Vert \Vert\vector{w}_j\Vert}{2\pi} \Big( \sin\gamma + (\pi - \gamma) \cos\gamma \Big),
\end{align*}
where $\gamma = \cos^{-1} \frac{\vector{w}_i^\top \vector{w}_j}{\Vert \vector{w}_i \Vert \Vert\vector{w}_j\Vert}$ is the angle between $\vector{w}_i$ and $\vector{w}_j$.
Using~\eqref{eq:gauss2sphere}, we may write the NNK on the sphere,
\begin{equation}
k_{(\cdot)^p\Theta(\cdot), \ident,U(\mathbb{S}^{d-1})}(\vector{\theta}_i, \vector{\theta}_j) = \frac{k_{(\cdot)^p\Theta(\cdot), \ident,\Phi}(\vector{\theta}_i, \vector{\theta}_j) \Gamma(d/2)}{2^p\Gamma(p+d/2)}, \label{eq:relu_sphere}
\end{equation}
where $\Gamma$ is the gamma function.
We use this newly derived kernel to build intensity functions on the sphere using neural networks with ReLU activations ($p=1$) in \S~\ref{sec:wildfires}.

\subsection{Extension to Product Spaces}
\label{sec:spatio-temporal}
It is sometimes helpful to view the domain $\mathbb{X}$ as a Cartesian product over multiple sets.
For example, we wish to model an intensity function that varies over the surface of the Earth as well as time, we might consider $\mathbb{X} = \mathbb{S}^2 \times \mathbb{T}$, where $\mathbb{S}^2$ is the unit sphere and $\mathbb{T}$ is some discrete or continuous set indexing time.
Closed-form integrated intensity functions then follow as a special case of the reasoning in earlier sections. 
A more detailed discussion of how to construct tractable product space intensities is given in Appendix~B.1.

\paragraph{Example}
We write $\mathbb{X} = \mathbb{Y} \times \mathbb{T}$, where $\mathbb{Y}$ might represent a spatial domain and $\mathbb{T}$ might represent a temporal domain.
We write $\vector{x} = (\vector{y}, \vector{\tau})$ and $\vector{\tau} \in \mathbb{T}$ and $\vector{y} \in \mathbb{Y}$. 
Let $\odot$ denote the Hadamard product and consider the model
\begin{align*}
    \lambda(\vector{x}) &= \alpha\big\Vert \matrix{V} \big( \vector{\psi_1}(\vector{y}) \odot \vector{\psi_2}(\vector{\tau}) \big) \big\Vert_2^2 \\
    &= \alpha \Tr\Big(\matrix{V}^\top \matrix{V} \big(\widetilde{\matrix{K}}_1(\vector{y})\odot \widetilde{\matrix{K}}_2(\vector{\tau}) \big) \Big), \quad \text{where}
\end{align*}
$
\widetilde{\matrix{K}}_1(\vector{y}) = \vector{\psi_1}({\vector{y}}) \vector{\psi_1}({\vector{y}})^\top$ and $ \widetilde{\matrix{K}}_2(\vector{\tau}) = \vector{\psi_2}({\vector{\tau}}) \vector{\psi_2}({\vector{\tau}})^\top.$
If the base measure $\mu$ decomposes as $\mu(\cdot) = \mu_1(\cdot) \mu_2(\cdot)$, the integrated intensity function is then
\begin{align*}
    \Lambda &= 
    \alpha \Tr\Big(\matrix{V}^\top \matrix{V} \big(\matrix{K}_1\odot \matrix{K}_2 \big) \Big),
\end{align*}
where the $ij$th entries of $\matrix{K}_1$ and $\matrix{K}_2$ are respectively
\begin{align*}
\int_{\mathbb{Y}}  \psi_{1i}({\vector{y}}) \psi_{1j}({\vector{y}}) d \mu_1(\vector{y}) \quad \text{and} \quad \int_{\mathbb{T}} \psi_{2i} ({\vector{\tau}}) \psi_{2j}({\vector{\tau}}) d\mu_2(\vector{\tau}).
\end{align*}
These kernels are tractable under the same settings~\eqref{eq:intensity1} and~\eqref{eq:intensity2}.
We consider the special case where $d=1$, $\vector{t}$ is the identity, $\vector{\psi}_2(\tau)=\text{ReLU}(w \tau + b)$, $\mu_2$ is Lebesgue and $\mathbb{T}=[T_1, T_2]$ is an interval in Appendix~C.

\section{Properties of SNEPPP}
\subsection{Optimisation Properties}
\label{sec:optimisation}
A common procedure for finding point estimates for $\lambda$ is via MLE.
A single realisation of a PPP observed over a window $A$ is $\{ \vector{x}_i \}_{i=1}^N$, where $N$ is itself a draw from a Poisson distribution according to~\eqref{eq:poisson}.
One may fit a PPP by observing a single realisation over $A$ and maximising the likelihood~\eqref{eq:joint} with respect to (the parameters of) $\lambda$, or alternatively by observing multiple independently drawn realisations over possibly different windows $A_1,\ldots,A_r$ and then maximising the product of likelihoods.
We focus here on the setting of a single realisation for notational simplicity and to follow previous works~\citep{flaxman2017poisson,walder2017fast}, but our discussion extends to multiple realisations.
It is likely that many of our insights extend beyond MLE (e.g. Bayesian inference schemes that utilise gradient-descent like algorithms). 
We assume that $\matrix{V}^\top \matrix{V}$ is positive definite, with smallest eigenvalue greater than or equal to $\epsilon_1>0$. 
That is, we set $\matrix{V}^\top \matrix{V}=\matrix{M} + \epsilon_1 \matrix{I}$ and reparameterise the model in terms of a PSD matrix $\matrix{M}$.

\paragraph{Convexity} Compare the two NLLs, under~\eqref{eq:conditional_points} and~\eqref{eq:joint} respectively, and note that the second NLL is strictly convex in $\matrix{M}$, being the sum of a linear penalty term and a strictly convex model fit term,
\begin{align*}
    &\phantom{{}={}}-\log p(\vector{x}_1,\ldots,\vector{x}_N, N|\lambda) =  \numberthis \label{eq:nll}\\
    &\quad \alpha \Tr \Big(  (\matrix{M} + \epsilon_1 \matrix{I}) \matrix{K}  \Big)  - \sum_{i=1}^N \log  \Tr \Big(  (\matrix{M} + \epsilon_1 \matrix{I}) \widetilde{\matrix{K}}(\vector{x}_i)  \Big) 
\end{align*}
In contrast, the NLL of~\eqref{eq:conditional_points}, which is the objective of an analogous density estimation procedure, is in general non-convex.
We expect that most easy to implement constrained optimisation routines perform well in optimising~\eqref{eq:nll} when the hidden layer is fixed. 
In fact, projected gradient descent (PGD) with sufficiently small constant step size $\eta$ converges.
PGD amounts to alternating gradient and projection iterations, beginning with initial guess $\matrix{M}_1$,
\begin{alignat*}{2}
    \matrix{M}_{t+1/2} &= \matrix{M}_t - \eta \nabla C(\matrix{M}_t) \quad &&\text{Gradient update} \\
    \matrix{M}_{t+1} &= \matrix{Q}_{t+1/2} \matrix{D}^+_{t+1/2} \matrix{Q}_{t+1/2}^\top \quad &&\text{Projection}, 
\end{alignat*}
where $\matrix{Q}\matrix{D}\matrix{Q}^\top$ is an eigendecomposition and $\matrix{D}^+$ is $\matrix{D}$ with negative entries set to zero\footnote{Frobenius norm projection of a symmetric matrix onto the space of PSD matrices sets the negative eigenvalues of the symmetric matrix to zero~\citep[\S~8.1.1]{boyd2004convex}.}.
We may optimise the likelihood or the posterior under a particular choice of prior. 
This prior is well-defined and amounts to a Gaussian distribution projected to the space of PSD matrices over $\matrix{M}$ --- see Appendix~A 
for more details.
\begin{restatable}{proposition}{optimisation}
    \label{prop:optimisation}
    Let $\mathbb{M}_+^{n} = \{ \matrix{M} \in \mathbb{R}^{n \times n } \mid \matrix{M} = \matrix{M}^\top, \matrix{M} \succeq 0\}$ denote the convex cone of PSD matrices. 
    Let $\epsilon_1 \geq 0$.
    The NLL~\eqref{eq:nll} is strictly convex in $\matrix{M}$ over $\mathbb{M}_+^{n }$. 
    Let $\epsilon_1 > 0$ and $\epsilon_2 \geq 0$, $\beta = \epsilon_2 + 1/\epsilon_1^2$ and consider the MLE or MAP estimate
    \begin{align*}
        \matrix{M}^\ast &= \argmin_{\matrix{M} \in \mathbb{M}_+^{n}} \overbrace{- \frac{1}{N}\log p(\vector{x}_1,\ldots,\vector{x}_N, N|\lambda) + \frac{\epsilon_2}{2} \Vert \matrix{M} \Vert_F^2}^{\triangleq C(\matrix{M})}.
    \end{align*}
    Choose the learning rate $\eta =1/\beta$. Then PGD satisfies
    \begin{align*}
        &\phantom{{}={}}C(\matrix{M}_{t+1}) - C(\matrix{M}^\ast) \leq \\
        &\frac{3\beta \Vert \matrix{M}_1 - \matrix{M}^\ast \Vert_F^2 + C(\matrix{M}_1) - C(\matrix{M}^\ast)}{t+1}.
    \end{align*}
    Suppose additionally that $\epsilon_2 > 0$. Then PGD satisfies
     \begin{align*}
         \Vert \matrix{M}_{t+1} - \matrix{M}^\ast \Vert_F^2 \leq \exp \Big( - \frac{t\epsilon_2}{2\beta} \Big) \Vert \matrix{M}_{1} - \matrix{M}^\ast \Vert_F^2.
     \end{align*}
\end{restatable}
It is straightforward take a Laplace approximation about the MAP. We leave this for future work.

\subsection{Special Cases}
\label{sec:special_cases}
\paragraph{Log-linear Models}
The log-linear model $\lambda(\vector{x}) = \exp(\vector{w}^\top \vector{x})$ is a widely used classical statistical model that is easy to integrate (see \S~\ref{sec:nnk_examples} for an example). 
We obtain this model as a special case of SNEPPP when $\sigma = \exp$, $m=n=1$ and the bias $\vector{b}$ is constrained to be zero. 
We refer to this model as ``log-linear'', and use it as a baseline for experiments in \S~\ref{sec:experiments}. 
Our implementation differs slightly to that of a typical statistical implementation; we use Adam to optimise parameters rather than second order methods.

\paragraph{Mixture Models} The model $\lambda(\vector{x}) = \sum_{i=1}^n v_{ii}^2 \exp(\vector{w}_i^\top \vector{x})$ is also easy to integrate, and is obtained as a special case of SNEPPP when $\sigma = \exp$, $m=n\geq 1$, the bias $\vector{b}$ is constrained to be zero, and the readout parameter $\matrix{V}$ is constrained to be diagonal. 
Borrowing terminology from i.i.d. density modelling, we refer to this model as ``log-linear mixture'', and use it as a baseline for experiments in \S~\ref{sec:experiments}.

\subsection{Related Work}
\label{sec:related}
\paragraph{Kernel Methods for Densities}
Our method uses the same machinery for closed-form integration as squared neural families (SNEFYs)~\citep{tsuchida2023squared} do for probability density models.
SNEFYs represent the unnormalised probability density as the squared norm of a single layer neural network.
SNEFYs in turn can be seen as neural network variants of kernel-based methods for representing probability densities~\citep{marteau2020non,rudi_ciliberto}.
These kernel methods amount to using something analogous to the square of a linear combination of elements of an RKHS for the unnormalised probability density.

\paragraph{Kernel Methods and Gaussian Processes for Intensities}
\citet{flaxman2017poisson} use a squared element of an RKHS as the intensity function.
A Bayesian extension is also available~\citep{walder2017fast}, where the intensity function is a squared Gaussian process (GP), and a Laplace approximation to the posterior is obtained.
An earlier work uses variational inference to obtain an approximate posterior under a squared GP model~\citep{lloyd2015variational}.
Random Fourier features are exploited by~\citep{sellier2023sparse} to build routines involving generalised stationary kernels.
We note that it would be relatively straightforward to compute Laplace approximations to the posterior by considering the Hessian about our MAP estimates. 
We leave empirical evaluation of such models for future work.

\paragraph{Optimisation Properties}
In all of the PPP works described above, the intensity function is a squared GP with a finite number of features, which is a special case of~\eqref{eq:intensity1} when $m=1$ under a Gaussian prior over $\matrix{V}$. 
To the best of our knowledge, we are the first to note that allowing $m = n$ and constraining $\matrix{V}^\top \matrix{V}$ to be positive definite allows for convex optimisation over $\matrix{V}^\top \matrix{V}$ for MLE or MAP estimates via Proposition~\ref{sec:optimisation}.
In previous works, optimisation over $\matrix{V}$ with $m=1$ is nonconvex (see for example~\citet[foonote 2]{flaxman2017poisson}).
Our new result is worth highlighting, given that in the density estimation setting for similar models, MLE or MAP estimates are also not convex, but other surrogate losses are used in their place~\citep[Theorem 7]{rudi_ciliberto}.
In contrast, our method allows for principled MLE and MAP estimation.

\paragraph{Augmented Permanental Point Processes}
\citet{kim2022fast} introduce augmented permanental point processes (APP), in which the intensity function is a squared GP of a function (called the covariate map) of the input variable $\vector{\rv{x}}$, rather than $\vector{\rv{x}}$ itself. 
The integrated intensity function is still computed with respect to $\vector{\rv{x}}$.
In the setting of APP, during training, the input $\vector{\rv{x}}$ and the covariate of $\vector{\rv{x}}$ are jointly observed, which differs to our setting in which only $\vector{\rv{x}}$ is observed.
Direct comparisons are therefore not possible, however we note that the implementation of~\citet{kim2022fast} are able to handle the non-augmented setting by choosing an identity covariate map.
In this setting, the resulting model is equivalent to that of~\citet{walder2017fast}, and the posterior mode is the squared RKHS model of~\citet{flaxman2017poisson}. 
The MAP optimisation problem involves a so-called equivalent kernel~\citep{williams2006gaussian,menon2018loss}, whose calculation can be performed in one of three methods: naive, random Fourier mappings (RFM)~\citep{rahimi2007random} and Nystr\"om~\citep{williams2000using}. 

\paragraph{Tractable Domains of Integration}
None of the related works on intensity estimation explicitly describe closed-form integration on interesting domains such as the hypersphere $\mathbb{S}^{d-1}$. 
In contrast, we may use the newly derived NNK~\eqref{eq:relu_sphere} or previously derived NNKs~\citep[Kernel 3]{tsuchida2023squared} to compute integrated intensity functions on the hypersphere.
\section{Experiments}
\label{sec:experiments}
\begin{table*}[t]
\centering
\fontsize{9}{10.8}
\selectfont
\begin{tabular}{c|c|c|c|c}
                   & NLL (exact)                    & NLL (MC)                       & Count percent error      & Time (seconds)   \\ \hline
Log-linear         & $-0.03 \pm 0.01$               & $-0.02 \pm 0.01$               & $\mathbf{0.03 \pm 0.03}$ & $1.79\pm 0.31$   \\
Log-linear mixture & $-0.07 \pm 0.02$               & $-0.07 \pm 0.02$               & $\mathbf{0.02 \pm 0.02}$ & $2.28 \pm 0.20$  \\
SNEPPP             & $\mathbf{-0.56 \pm 0.01^\ast}$ & $\mathbf{-0.62 \pm 0.01^\ast}$ & $\mathbf{0.04 \pm 0.03}$ & $6.20 \pm 0.18$  \\
RFM                & n/a                            & $-0.38 \pm 0.03$               & $0.07 \pm 0.03$          & $7.80 \pm 0.73$  \\
Nystr\"om            & n/a                            & $-0.52 \pm 0.02$               & $0.09 \pm 0.03$          & $4.74 \pm 0.33$  \\
Naive              & n/a                            & $-0.51 \pm 0.03$               & $0.09 \pm 0.04$          & $18.48 \pm 0.48$ \\
\hline \hline
Log-linear         & $-0.04 \pm 0.01$               & $-0.04 \pm 0.01$          & $0.04 \pm 0.02$          & $2.38\pm 0.14$   \\
Log-linear mixture & $-0.08 \pm 0.01$               & $-0.08 \pm 0.01$          & $\mathbf{0.03 \pm 0.02}$ & $2.45 \pm 0.18$  \\
SNEPPP             & $\mathbf{-0.78 \pm 0.03^\ast}$ & $\mathbf{-0.84 \pm 0.03}$ & $\mathbf{0.03 \pm 0.02}$ & $7.07 \pm 0.17$  \\
RFM                & n/a                            & $-0.23 \pm 0.02$          & $0.07 \pm 0.03$          & $4.17 \pm 0.31$  \\
Nystr\"om            & n/a                            & $-0.35 \pm 0.03$          & $0.09 \pm 0.04$          & $4.92 \pm 0.48$  \\
Naive              & n/a                            & $-0.36 \pm 0.03$          & $0.11 \pm 0.05$          & $25.88 \pm 1.10$
\end{tabular}
\caption{ Shown are means $\pm$ sample standard deviations over $100$ randomly generated synthetic datasets and parameter initialisations. Asterisked and bold values are significantly better than closest competitor according to a two sample t-test. Bold numbers indicate values which are not significantly different to the best. Some methods do not provide exact integrated intensity functions or NLL, so we resort to Monte Carlo (MC) estimation. (Top) Bei real data experiment results.\label{tab:bei} (Bottom) clmfires real data experiment results.\label{tab:clmfires} Results for the Copper dataset are given in 
Table~3, Appendix~D.}
\end{table*}

\subsection{Metrics for Intensity Functions}
There are various ways in which one may asses the quality of an intensity function $\lambda$ with respect to held-out test data $\{ \tilde{\vector{x}}_i \}_{i=1}^M$ observed over a potentially partial window $A \subseteq \mathbb{X}$.
There are some important distinctions between assessing the quality of density models~\eqref{eq:conditional_points} and PPP models~\eqref{eq:joint}.
We briefly describe three ways in which PPPs can be evaluated.

\paragraph{Test Log Likelihood}
We evaluate the test log likelihood on a held out piece of data that is independent of the data used for training in two ways.
Firstly, we may exploit the \emph{complete independence of disjoint subsets} of PPPs. Suppose for example that we observe and fit an intensity function to wildfires observed over the entire Earth for years $2003$ to $2023$, excluding the year $2018$. Each of $\rv{N}\big(\mathbb{S}^2 \times \{ \tau\}\big)$ for distinct $\tau \in \{ 2003, \ldots, 2023\}$ are mutually independent. 
We may therefore test on data from $2018$. 
The same idea applies to spatially disjoint observations.
Secondly, we may artificially but exactly obtain multiple independent realisations from a single realisation of data by a process called \emph{splitting or thinning}, see Appendix~D. 

\paragraph{Predictions of Counts} Suppose we fit a PPP using data from a window that does not include $A \subseteq \mathbb{X}$. 
We reserve observations over the window $A$ for testing.
We predict the number of events within $A$ and compare the result with such test observations. 
Alternatively, we may predict the number of events in any window and compare the result with an independent realisation over the same window.

\paragraph{Ground Truth Intensities} A final option is to compare the fitted intensity function $\lambda$ against a ground truth intensity function $\lambda_{\text{GT}}$.
Such comparisons can only be made when using synthetic data.
We consider the root mean squared error (RMSE) over a random sample $\{ \vector{z}_i\}_{i=1}^P$ taken uniformly from $\mathbb{X}$, $\text{RMSE}^2 = \frac{1}{P} \sum_{i=1}^P \big(\lambda(\vector{z}_i) - \lambda_{\text{GT}} (\vector{z}_i) \big)^2 \approx \frac{1}{\big| \mathbb{X}\big| }\int_{\mathbb{X}} \big( \lambda(\vector{x}) - \lambda_{\text{GT}} (\vector{x}) \big)^2 d\vector{x}$.

\subsection{Baselines}
\paragraph{Log-linear and Mixture Models}
We compare against log-linear and log-mixture models, which are implemented as a special case of SNEPPP (see \S~\ref{sec:special_cases}).

\paragraph{Kernel and GP Methods}
We use an open source TensorFlow library~\citep{kim2022fast} for GP baselines that compute the equivalent kernel using Naive, Nystr\"om and Random Fourier Methods (RFM).
The computational complexity scales cubically in the number of data points in the Naive case, and cubically in the number of features in the Nystr\"om or RFM case.
While exact NLL is in theory possible using the RFM and Nystr\"om methods using the same techniques we discuss here, they are not implemented in the software package that we use.
We do not compare with APP which covers a different problem setting (see \S~\ref{sec:related}).

\begin{figure*}
    \centering
    \includegraphics[scale=0.19]{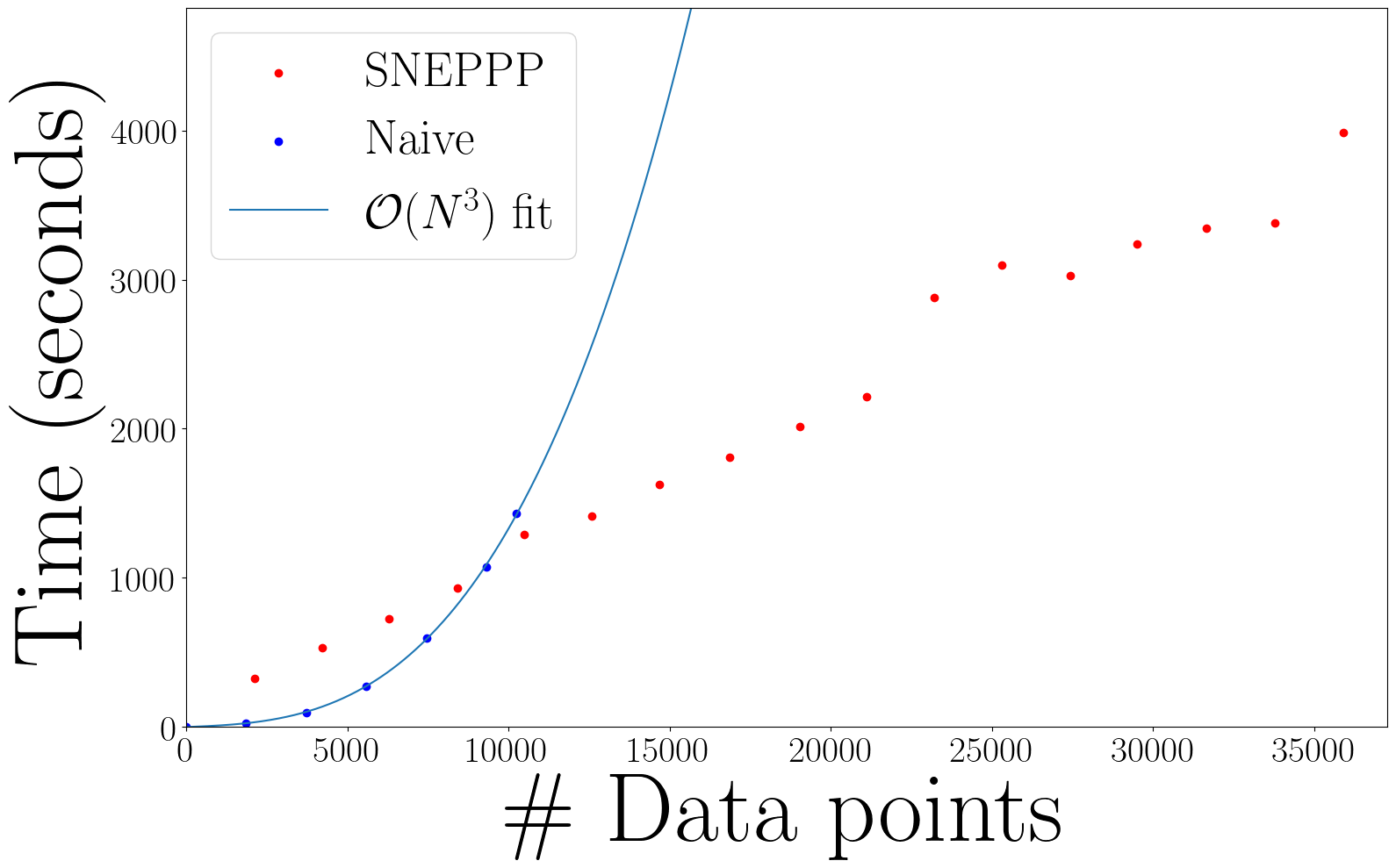} \includegraphics[scale=0.19]{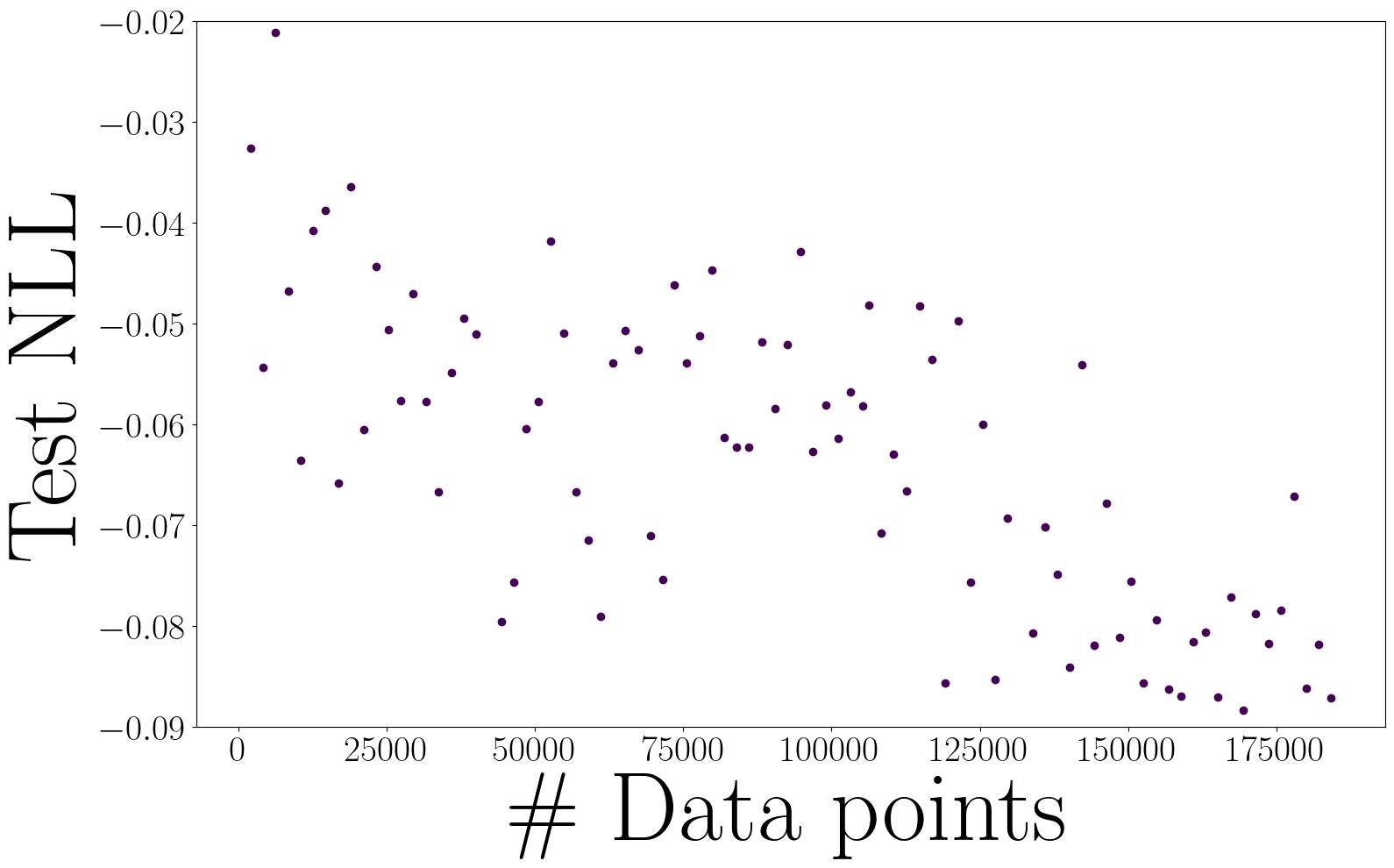}
    \caption{(Left) Using a naive kernel method or squared GP requires solving a linear system, incurring $\mathcal{O}(N^3)$ time. In contrast, finite feature methods, such as SNEPPP, scale linearly in the number of datapoints. In this setting, the naive method runs out of memory beyond $5.5\%$ of data usage, but even if we had sufficient memory to perform the required calculation, it would take roughly $88$ days to compute using $100\%$ data. In contrast, SNEPPP takes less than 8 hours. (Right) Training SNEPPPs using more data decreases the test NLL, after appropriately accounting for thinning. }
    \label{fig:spatial_time}
\end{figure*}

\subsection{Evaluation on Synthetic and Real Data}
We measure the NLL (using a thinning process), RMSE and computation time (in seconds) for an existing synthetic setting~\citep{baddeley2012nonparametric,kim2022fast}.
Full details and results are given in Appendix~D. 
We also perform  benchmarks on three real datasets --- bei, copper and clmfires --- also considered by~\citet{kim2022fast}. 
See Appendix~D 
 for details.
Results are summarised in Tables~\ref{tab:bei} and~3.

\subsection{Case Study on Wildfires}
\label{sec:wildfires}
We now turn our attention to the spatio-temporal modelling of the incidence of wildfires around planet Earth.
We use a massive freely available dataset~\citep{bushfires}.
In the November and December months of the year 2000, over 200,000 datapoints are observed --- we call this restricted dataset the spatial dataset.
Between the years 2000 and 2022 over 98 million datapoints are observed --- we call this complete dataset the spatio-temporal dataset.
In this setting, having an observation domain over the sphere (and potentially the time domain) and a huge number of events, we are not aware of any publically available software tools that we can quantitatively compare with. 
Instead, we highlight attractive features of applying our model to big data, and compare performance against the special cases described in \S~\ref{sec:special_cases}.

\paragraph{Finite Feature but not Naive Methods Scale to Big Data}
We apply thinning to the process and compare the training time for a SNEPPP model and a Naive permanental point process model on the spatial dataset.
We use the squared exponential kernel implemented by~\citet{kim2022fast}, which assumes a domain of $\mathbb{X}=\mathbb{R}^d$ instead of $\mathbb{X}=\mathbb{S}^{d-1}$, noting that we are only interested in computation time.
Such a model stands in as a good proxy in terms of computation time for any future developments of naive equivalent kernels for $\mathbb{X}=\mathbb{S}^{d-1}$.
We find that the SNEPPP model is able to use all of the data, and its training time scales favourably with the number of training points.
In contrast, the naive model is too memory intensive to use all data, and its training time scales poorly with the number of training points.
See Figure~\ref{fig:spatial_time}.

\paragraph{Using More Data Improves Predictive Performance}
We measure the predictive performance of SNEPPP as the number of training examples increases in Figure~\ref{fig:spatial_time}.
Note that events are not sampled i.i.d. from some distribution and we do not necessarily expect for standard generalisation error bounds to apply, even for fixed feature models such as~\citet[Theorem 5]{marteau2020non}.
While we do not necessarily expect the generalisation gap to decay like $\mathcal{O}(N^{-1/2})$ as in the i.i.d. setting, we empirically find that both the test NLL and the error in the number of predicted events improves as more training data is added.
We leave theoretical analysis for future work.

\paragraph{Fit to 100 Million Spatio-temporal Events}
We use a spatio-temporal model with the temporal component as described in \S~\ref{sec:spatio-temporal} and a spatial component consisting of ReLU activations according to~\eqref{eq:relu_sphere}.
We provide a video of our fitted intensity functions varying over time in the supplementary material and Appendix~D. 
Due to the linear scaling in $N$ of our closed form integral in SNEPPP, 
this result only required 8 hours of computation time for fitting.
We also find that the general SNEPPP model shows increased performance against (mixtures of) log-linear models.

\section{Conclusion}
We propose an efficient and flexible model for PPPs based on two-layer squared neural networks. 
In contrast to recent work~\citep{tsuchida2023squared}
which considered density estimation, this paper uses the neural network to model the expected value (i.e. intensity) of a PPP.
We prove that the resulting integrated intensity function has a closed form.
Based on the link between neural networks and kernel methods, this closed form provides a linear time (with respect to input dimension) computation of the integrated intensity function.
We also derive two novel NNKs which are well-suited to fitting
a large scale wildfire model.
We derive a projected gradient descent approach for optimising the NLL, and show that the problem is convex if we hold the first neural network layer fixed.
We illustrate the efficacy of our method with empirical experiments on several different performance metrics.
Our empirical results show that our exact integration method has similar computation time requirements, but outperform other baseline methods.
As a demonstration of the efficiency of our approach, we apply SNEPPP to a case study of 100 million wildfire events. 

\paragraph{Limitations and Future Work} In order to simplify our notation, we assumed that a single activation function $\sigma$ is used.
One may straightforwardly extend our description to the case where more than one activation function is used in the same layer, as might be the case for certain random Fourier feature inspired feature mappings with lower variance than their single activation function counterparts~\citep{sutherland2015error}.
Our empirical results indicate that increasing the size of the training data improves test likelihood. While the generalisation error of supervised learning is theoretically well studied, we are unaware of any rates of convergence of generalisation error for PPP. 
\section*{Acknowledgements}
Russell and Cheng Soon would like to acknowledge the support of the Machine Learning and
Artificial Intelligence Future Science Platform, CSIRO. The authors would like to thank Jia Liu for
early discussions about the idea.
\bibliography{refs}

\onecolumn
\appendix
\section*{Appendix}
\section{Proofs}
\label{sec:proofs}
\trace*
\begin{proof}
The intensity function $\lambda$ is quadratic in $\matrix{V}$.
More specifically, we have
\begin{align*}
    \lambda(\vector{x}) &= \alpha\big\Vert \matrix{V} \vector{\psi}(\vector{x})\big\Vert_2^2 \\
    &= \alpha \vector{\psi}^\top \matrix{V}^\top \matrix{V} \vector{\psi} \\
    &= \alpha \Tr \Big( \vector{\psi}^\top \matrix{V}^\top \matrix{V} \vector{\psi}  \Big) \\ 
    &= \alpha \Tr \Big(  \matrix{V}^\top \matrix{V} \vector{\psi} \vector{\psi}^\top  \Big).
\end{align*}
Using this quadratic representation and linearity of the trace, we have
\begin{align*}
    \Lambda &= \int_{\mathbb{X}} \lambda(\vector{x}) \mu(d\vector{x}) \\
    &=\alpha  \int_{\mathbb{X}} \Tr\Big(\matrix{V}^\top \matrix{V}  \vector{\psi} \vector{\psi}^\top   \Big) \mu(d\vector{x}) \\ 
    &=\alpha  \Tr\Big(\matrix{V}^\top \matrix{V}  \int_{\mathbb{X}}  \vector{\psi} \vector{\psi}^\top  \mu(d\vector{x})  \Big) \\ 
    &= \alpha \Tr\Big(\matrix{V}^\top \matrix{V} \matrix{K}\Big),
\end{align*}
where $\matrix{K}$ is the PSD matrix with $ij$th entry $\kappa(i,j)$. 
\end{proof}

\paragraph{Prior construction}
We place a prior over $\matrix{M}$ which is proportional to $\exp(-\frac{\epsilon_1}{2} \Vert \matrix{M} \Vert_F^2)$ and supported on the space of PSD matrices. 
The normalising constant, $\int_{\mathbb{M}_+^n} \exp(-\frac{\epsilon_1}{2} \Vert \matrix{M} \Vert_F^2) \, d\matrix{M}$, is nonzero and finite, being bounded below by zero and bounded above by the normalising constant of an isotropic Gaussian distribution.

\optimisation*
\begin{proof}
    The negative log likelihood $- \log p(\vector{x}_1,\ldots,\vector{x}_N, N|\lambda)$ is strictly convex in $\matrix{M}$ by linearity of the trace and strict convexity of $-\log(\cdot)$. 
    
    The gradient of the objective is given by
    \begin{align*}
    \nabla C = \frac{\partial C}{\partial \matrix{M}} = \frac{1}{N}\alpha \matrix{K} - \frac{1}{N}\sum_{i=1}^N \frac{\widetilde{\matrix{K}}(\vector{x}_i)}{\Tr\big((\matrix{M} + \epsilon_1 I )\widetilde{\matrix{K}}(\vector{x}_i) \big)} + \epsilon_2 \matrix{M}.
    \end{align*}
    The gradient norm satisfies
    \begin{align*}
       &\phantom{{}={}} \big\Vert \nabla C(\matrix{M}_1) - \nabla C (\matrix{M}_2) \big\Vert_F \\
       &\leq \epsilon_2  \Vert \matrix{M}_1 - \matrix{M}_2 \Vert_F + \frac{1}{N}\sum_{i=1}^N \Bigg\Vert \frac{\widetilde{\matrix{K}}(\vector{x}_i)}{\Tr\big((\matrix{M}_1 + \epsilon_1 I )\widetilde{\matrix{K}}(\vector{x}_i) \big)} - \frac{\widetilde{\matrix{K}}(\vector{x}_i)}{\Tr\big((\matrix{M}_2 + \epsilon_1 I )\widetilde{\matrix{K}}(\vector{x}_i) \big)} \Bigg\Vert_F \\
       &= \epsilon_2  \Vert \matrix{M}_1 - \matrix{M}_2 \Vert_F + \frac{1}{N}\sum_{i=1}^N \Bigg\Vert \frac{\widetilde{\matrix{K}}(\vector{x}_i)\Big( \Tr\big((\matrix{M}_2 + \epsilon_1 I )\widetilde{\matrix{K}}(\vector{x}_i) \big) - \Tr\big((\matrix{M}_1 + \epsilon_1 I )\widetilde{\matrix{K}}(\vector{x}_i) \big)}{\Tr\big((\matrix{M}_1 + \epsilon_1 I )\widetilde{\matrix{K}}(\vector{x}_i) \big)\Tr\big((\matrix{M}_2 + \epsilon_1 I )\widetilde{\matrix{K}}(\vector{x}_i) \big)} \Bigg\Vert_F \\
       &\leq \epsilon_2  \Vert \matrix{M}_1 - \matrix{M}_2 \Vert_F +  \frac{1}{N}\sum_{i=1}^N\Bigg\Vert \frac{\widetilde{\matrix{K}}(\vector{x}_i)}{\epsilon_1^2 \Tr\big(\widetilde{\matrix{K}}(\vector{x}_i) \big)^2} \Bigg\Vert_F \Big| \Tr\big((\matrix{M}_2 - \matrix{M}_1 )\widetilde{\matrix{K}}(\vector{x}_i) \big) \Big| \\ 
       &\leq \epsilon_2  \Vert \matrix{M}_1 - \matrix{M}_2 \Vert_F +  \frac{1}{N \epsilon_1^2}\sum_{i=1}^N \Big( \frac{\Vert \widetilde{\matrix{K}}(\vector{x}_i) \Vert_F }{ \Tr\big(\widetilde{\matrix{K}}(\vector{x}_i) \big)} \Big)^2  \Big\Vert \matrix{M}_2 - \matrix{M}_1  \Big\Vert_F \\ 
       &= \Big(\epsilon_2 + 1/\epsilon_1^2 \Big) \Vert \matrix{M}_1 - \matrix{M}_2 \Vert_F,
    \end{align*}
where the last line follows because $\widetilde{\matrix{K}}(\vector{x}_i)$ is rank $1$, so its sum of squared eigenvalues is equal to the sum of its eigenvalues squared.
The gradient is therefore $\beta$-Lipschitz, where $\beta = (\epsilon_2 + N/\epsilon_1^2)$. 
Theorem 3.7 of~\citet{bubeck2015convex} then gives the required result under convexity and $\beta$-Lipschitzness, and Theorem 3.10 of~\citet{bubeck2015convex} gives the required result under strong convexity and $\beta$-Lipschitzness.
\end{proof}

\section{Extension to product spaces}
\label{sec:product}
In certain settings, it is natural to consider $\mathbb{X}$ as being a Cartesian product over multiple sets. 
For example, we might wish to model an intensity function that varies over the surface of the Earth and in time.
In such settings, it is useful to consider a slight extension of the models~\eqref{eq:intensity1} and~\eqref{eq:intensity2}. 
The mechanics of this extension follow naturally from the previously presented models; here we provide the details.
We write $\mathbb{X} = \mathbb{Y} \times \mathbb{T}$, where $\mathbb{Y}$ might represent a spatial domain and $\mathbb{T}$ might represent a temporal domain, but this need not be the case.
We write $\vector{x} = (\vector{y}, \vector{\tau})$ and $\vector{\tau} \in \mathbb{T}$ and $\vector{y} \in \mathbb{Y}$. 
Let $\odot$ denote the Hadamard product and consider the model
\begin{align*}
    \lambda(\vector{x}) = \alpha\big\Vert \matrix{V} \big( \vector{\psi_1}(\vector{y}) \odot \vector{\psi_2}(\vector{\tau}) \big) \big\Vert_2^2 &= \alpha \Tr\Big(\matrix{V}^\top \matrix{V} \big(\vector{\psi_1}(\vector{y})  \odot \vector{\psi_2}(\vector{\tau}) \big) \big(\vector{\psi_1}(\vector{y})  \odot \vector{\psi_2}(\vector{\tau}) \big) ^\top \Big) , \numberthis \label{eq:intensity3} \\
    &= \alpha \Tr\Big(\matrix{V}^\top \matrix{V} \big(\widetilde{\matrix{K}}_1(\vector{y})\odot \widetilde{\matrix{K}}_2(\vector{\tau}) \big) \Big),
\end{align*}
where
\begin{align*}
\widetilde{\matrix{K}}_1(\vector{y}) = \vector{\psi_1}({\vector{y}}) \vector{\psi_1}({\vector{y}})^\top \quad \text{and} \quad \widetilde{\matrix{K}}_2(\vector{\tau}) = \vector{\psi_2}({\vector{\tau}}) \vector{\psi_2}({\vector{\tau}})^\top.
\end{align*}
If the base measure $\mu$ decomposes as $\mu(\cdot) = \mu_1(\cdot) \mu_2(\cdot)$ over sigma algebras generated by $\mathbb{Y}$ and $\mathbb{T}$, the integrated intensity function is then
\begin{align*}
    \Lambda &= \int_{\mathbb{T}} \int_{\mathbb{Y}} \alpha\big\Vert \matrix{V} \big( \vector{\psi_1}(\vector{y}) \odot \vector{\psi_2}(\vector{\tau}) \big) \big\Vert_2^2 \, d\mu_1(\vector{y}) d\mu_2(\vector{\tau}) \\
    &=\alpha \Tr\Big(\matrix{V}^\top \matrix{V} \big(\matrix{K}_1\odot \matrix{K}_2 \big) \Big),
\end{align*}
where the $ij$th entries of $\matrix{K}_1$ and $\matrix{K}_2$ are respectively
\begin{align*}
\int_{\mathbb{Y}}  \psi_{1i}({\vector{y}}) \psi_{1j}({\vector{y}}) d \mu_1(\vector{y}) \quad \text{and} \quad \int_{\mathbb{T}} \psi_{2i} ({\vector{\tau}}) \psi_{2j}({\vector{\tau}}) d\mu_2(\vector{\tau}).
\end{align*}
These kernels are then tractable under the same settings~\eqref{eq:intensity1} and~\eqref{eq:intensity2} as discussed in the main paper. 

\paragraph{Disambiguating separability} Note that in kernel literature, the kernel $\matrix{K}_1 \odot \matrix{K}_2$ has entries which are evaluations of a so-called \emph{separable kernel}.
This should not be confused with the use of the word separable in reference to the intensity function. Recall that our kernels are defined on the parameter space of the individual feature mappings $\vector{\psi_1}$ and $\vector{\psi_2}$, not on domains $\mathbb{Y}$ and $\mathbb{T}$.
The resulting intensity function under~\eqref{eq:intensity3} is not in general a \emph{separable intensity function}, which would be an intensity function satisfying $\lambda(\vector{y}, \vector{t}) = \lambda(\vector{y})\lambda(\vector{t})$ since the final layer ``mixes'' the individual feature mappings $\vector{\psi_1}$ and $\vector{\psi_2}$.

\subsection{Tractable product space constructions: a more general view}
\label{sec:product-space-general}
More generally, we may consider other ways of dealing with product spaces. 
We discuss several approaches in this sub-appendix, and leave their empirical evaluation for future work.

When considering $\mathbb{X} = \mathbb{Y} \times \mathbb{T}$ where $\mathbb{Y}$ and $\mathbb{T}$ are two different domains, we have several different ways to approach the construction of a tractable density or intensity model on $\mathbb{X}$. The main distinction is the level of the neural network architecture at which the inputs $\vector{y}$ and $\vector{\tau}$ start ``mixing'', i.e. sharing the parameters. Intuitively, if they mixing at a lower level of the architecture, we capture richer dependence structure between $\vector{y}$ and $\vector{\tau}$, but we also want to ensure that the model is still tractable, so it may be sensible to delay parameter sharing to a higher level in order to maintain analytical tractability. Like before, we assume a product base measure $\mu(\cdot) = \mu_1(\cdot) \mu_2(\cdot)$.
\begin{itemize}
    \item \textbf{At the input level}: Joint warping function $\vector{t}(\vector{y},\vector{\tau})$ -- this ``mixes'' $\vector{y}$ and $\vector{\tau}$ at the lowest level, but it is unlikely to lead to a tractable model in any interesting cases when we integrate with respect to a product base measure $\mu$.
    \item \textbf{At the first layer}: Use $\vector{t}(\vector{y},\vector{\tau})=\left[\begin{array}{c}
\vector{t_1}(\vector{y})
\\\vector{t_2}(\vector{\tau})
\end{array}\right]$, and thus 
$\lambda(\vector{x}) = \alpha\big\Vert \matrix{V} \vector{\sigma}\big( \matrix{W_1}\vector{t_1}(\vector{y})+ \matrix{W_2}\vector{t_2}(\vector{\tau})+\vector{b} \big) \big\Vert_2^2$.
This is analogous to the joint SNEFY density model in \citet{tsuchida2023squared} which satisfies desirable properties like being closed under conditioning. If $\mathbb{Y}$ and $\mathbb{T}$ are both Euclidean spaces, there is a number of options which are tractable. However, it may be difficult to obtain tractability of the joint model if the domains $\mathbb{Y}$ and $\mathbb{T}$ are of a very different nature, e.g. sphere and a real line, as one needs to match the activation $\sigma$ with the appropriate base measure $\mu$.
\item \textbf{At the second layer}: Keep the feature mappings $\vector{\psi_1}(\vector{y})=\vector{\sigma_1}\big( \matrix{W_1}\vector{t_1}(\vector{y}) +\vector{b_1}\big) $ and $\vector{\psi_2}(\vector{\tau})=\vector{\sigma_2}\big( \matrix{W_2}\vector{t_2}(\vector{\tau})+\vector{b_2} \big) $ arising in the first layer separate (note that they can even have different activations) and combine them as inputs to the second layer. We identify two approaches to combining them that \emph{always} lead to tractable product models if the combinations of $\sigma_1$ and $\mu_1$ and $\sigma_2$ and $\mu_2$ both lead to tractable individual models on $\mathbb{Y}$ and $\mathbb{T}$ respectively.
\begin{enumerate}
    \item \textit{Entrywise product}: Assuming $\vector{\psi_1}(\vector{y})$ and $\vector{\psi_2}(\vector{\tau})$ have the same dimension $n$, take $\vector{\psi}(\vector{y},\vector{\tau})=\vector{\psi_1}(\vector{y}) \odot \vector{\psi_2}(\vector{\tau})$ and use $\matrix{V}\in\mathbb R^{m\times n}$ as before. This leads to $\matrix{K}=\matrix{K}_1 \odot \matrix{K}_2$ (Hadamard product).
    \item \textit{Outer product}: Let $\vector{\psi_1}(\vector{y})\in\mathbb R^{n_1}$ and $\vector{\psi_2}(\vector{\tau})\in\mathbb R^{n_2}$ and take $\vector{\psi}(\vector{y},\vector{\tau})=\vector{\psi_1}(\vector{y})\vector{\psi_2}(\vector{\tau})^\top$. The width of the first layer is now $n_1n_2$ so $\matrix{V}\in\mathbb R^{m\times (n_1n_2)}$. This leads to $\matrix{K}=\matrix{K}_1 \otimes \matrix{K}_2$ (Kronecker product). While always analytically tractable, the higher number of parameters and computational cost may limit the practical utility of this approach.
\end{enumerate}
If the integrals of the form 
$$c_1(i)=\int \sigma(\vector{w_{1i}}^\top \vector{t_1}(\vector{y})+b_{1i})\mu_1(d\vector{y}),\quad c_2(i)=\int \sigma(\vector{w_{2i}}^\top \vector{t_2}(\vector{\tau})+b_{2i})\mu_2(d\vector{\tau})$$
are also tractable on the individual domains (note that these are generally simpler then the ones involving the products of $\sigma$ terms), we have two more options. Note that both of these options result in functions that include interaction terms so are not additive in $\vector{y}$ and $\vector{\tau}$.
\begin{enumerate}
    \item[3.] \textit{Addition}: Assuming $\vector{\psi_1}(\vector{y})$ and $\vector{\psi_2}(\vector{\tau})$ have the same dimension $n$, take $\vector{\psi}(\vector{y},\vector{\tau})=\vector{\psi_1}(\vector{y})+\vector{\psi_2}(\vector{\tau})$ and use $\matrix{V}\in\mathbb R^{m\times n}$ as usual. This leads to
        \begin{eqnarray*}
        \lambda(\vector{y}, \vector{t}) &=& \alpha\left(\big\Vert  \matrix{V} \vector{\psi_1} \big\Vert_2^2+\big\Vert  \matrix{V} \vector{\psi_2}  \big\Vert_2^2+\Tr(\matrix{V}^\top \matrix{V}(\vector{\psi_1}\vector{\psi_2}^\top+\vector{\psi_2}\vector{\psi_1}^\top))\right),
        \\\Lambda&=&\alpha\left(\Tr\Big(\matrix{V}^\top \matrix{V} \matrix{K_1} \Big)+\Tr\Big(\matrix{V}^\top \matrix{V} \matrix{K_2} \Big)+\Tr(\matrix{V}^\top \matrix{V}(\vector{c_1}\vector{c_2}^\top+\vector{c_2}\vector{c_1}^\top))\right).
    \end{eqnarray*}
    \item[4.] \textit{Stacking}: Let $\vector{\psi_1}(\vector{y})\in\mathbb R^{n_1}$ and $\vector{\psi_2}(\vector{\tau})\in\mathbb R^{n_2}$ and take $\vector{\psi}(\vector{y},\vector{\tau})=\left[\begin{array}{c}
\vector{\psi_1}(\vector{y})
\\\vector{\psi_2}(\vector{\tau})
\end{array}\right]$. The width of the first layer is now $n_1+n_2$ so take $\matrix{V}=[\matrix{V_1}\,\matrix{V_2}]\in\mathbb R^{m\times (n_1+n_2)}$. Here
        \begin{eqnarray*}
        \lambda(\vector{y}, \vector{t}) &=& \alpha\left(\big\Vert  \matrix{V_1} \vector{\psi_1} \big\Vert_2^2+\big\Vert  \matrix{V_2} \vector{\psi_2}  \big\Vert_2^2+\Tr(\matrix{V_2}^\top \matrix{V_1}\vector{\psi_1}\vector{\psi_2}^\top+\matrix{V_1}^\top \matrix{V_2}\vector{\psi_2}\vector{\psi_1}^\top)\right),
        \\\Lambda&=&\alpha\left(\Tr\Big(\matrix{V_1}^\top \matrix{V_1} \matrix{K_1} \Big)+\Tr\Big(\matrix{V_2}^\top \matrix{V_2} \matrix{K_2} \Big)+\Tr(\matrix{V_2}^\top \matrix{V_1}\vector{c_1}\vector{c_2}^\top+\matrix{V_1}^\top \matrix{V_2}\vector{c_2}\vector{c_1}^\top)\right).
    \end{eqnarray*}
\end{enumerate}
\item \textbf{At the output level}: We can simply combine two intensity models, either in an additive $\lambda(\vector{y}, \vector{t}) = \lambda(\vector{y})+\lambda(\vector{t})$, or separable fashion $\lambda(\vector{y}, \vector{t}) = \lambda(\vector{y})\lambda(\vector{t})$. Tractability is immediate, but the model may be too simplistic in many cases. A separable model in fact also arises from the outer product ``mixing'' where $\vector{\psi}(\vector{y},\vector{\tau})=\vector{\psi_1}(\vector{y})\vector{\psi_2}(\vector{\tau})^\top$ but we use the parametrisation $\matrix{V}=\matrix{V_1}\otimes \matrix{V_2}$. In this case we obtain
    \begin{eqnarray*}
        \lambda(\vector{y}, \vector{t}) &=& \alpha\big\Vert ( \matrix{V_1} \vector{\psi_1} ) \otimes ( \matrix{V_2} \vector{\psi_2} ) \big\Vert_2^2=\alpha\big\Vert  \matrix{V_1} \vector{\psi_1} \big\Vert_2^2\big\Vert  \matrix{V_2} \vector{\psi_2}  \big\Vert_2^2,
        \\\Lambda&=&\alpha\Tr( (\matrix{V_1}^\top\matrix{V_1} \matrix{K_1} )\otimes (\matrix{V_2}^\top\matrix{V_2} \matrix{K_2} ))=\alpha\Tr( \matrix{V_1}^\top\matrix{V_1} \matrix{K_1} ) \Tr(\matrix{V_2}^\top\matrix{V_2} \matrix{K_2} ).
    \end{eqnarray*}
Hence, attempting to make the outer product mixing more scalable using Kronecker algebra will significantly reduce its expressivity.

\end{itemize}
We conclude that mixing at the second layer with an entrywise product strikes the right balance between tractability, expressivity, and computational cost.

\section{A new NNK with ReLU activations}
\label{app:relu1dNNK}
In general, the NNK with ReLU activations, Gaussian base measure and identity sufficient statistic does not admit a closed form when $\vector{b} \neq \vector{0}$.
However, in the case that $d=1$ and the base measure is Lebesgue on some interval, the NNK does admit a closed form. 
The NNK is given by
\begin{align*}
    k_{\text{ReLU}, \ident, d\vector{x}}(\vector{\theta}_1, \vector{\theta}_2) &= \int_{T_1}^{T_2} \text{ReLU}(w_1 \tau + b_1) \text{ReLU}(w_2 \tau + b_2) d\tau, \quad T_2 > T_1 \\
    &= \int_{-\infty}^{\infty} (w_1 \tau + b_1) (w_2 \tau + b_2) \Big( \Theta( \tau - T_1) \Theta( T_2 - \tau) \Theta(w_1 \tau + b_1) \Theta(w_2 \tau + b_2) \Big) d\tau.
\end{align*}
The product of Heaviside step functions is one inside an interval $(a_1, a_2)$ and zero outside the interval.
The bounds of the interval take four values, depending on $b_1/w_1$ and $b_2/w_2$, and are the intersection of the regions $\tau > T_1$, $\tau < T_2$, $w_1 \tau + b_1 > 0$ and $w_2 \tau + b_2 > 0$. 
These inequalities can be rearranged to obtain the values of $a_1$ and $a_2$, depending on the signs of $w_1$ and $w_2$, as follows.
\begin{alignat*}{4}
    w_1 > 0, w_2 &> 0: \qquad a_1 &&= \max\Big(T_1, -\frac{b_1}{w_1}, -\frac{b_2}{w_2} \Big)  \qquad && a_2 &&= \max(a_1, T_2) \\
    w_1 < 0, w_2 &> 0: \qquad a_1 &&= \max\Big(T_1, -\frac{b_2}{w_2} \Big)  \qquad && a_2 &&= \max\Bigg(a_1, \min\Big(T_2, -\frac{b_1}{w_1} \Big) \Bigg) \\
    w_1 > 0, w_2 &< 0: \qquad a_1 &&= \max\Big(T_1, -\frac{b_1}{w_1} \Big)  \qquad && a_2 &&= \max\Bigg( a_1, \min\Big(T_2, -\frac{b_2}{w_2} \Big) \Bigg) \\
    w_1 < 0, w_2 &< 0: \qquad a_1 &&= T_1  \qquad && a_2 &&= \max\Bigg(a_1, \min\Big(T_2, -\frac{b_1}{w_1}, -\frac{b_2}{w_2} \Big) \Bigg).
\end{alignat*}
We then have
\begin{align*}
    k_{\text{ReLU}, \ident, d\vector{x}}(\vector{\theta}_1, \vector{\theta}_2) &= \int_{a_1}^{a_2} w_1 w_2 \tau^2 + (w_1 b_2 + w_2 b_1) \tau + b_1 b_2 \, d\tau \\
    &= \Big( \frac{w_1 w_2}{3} \tau^3 + \frac{w_1 b_2 + w_2 b_1}{2} \tau^2 + b_1 b_2 \tau \Big)\Big|_{\tau=a_1}^{\tau=a_2}
\end{align*}

\section{Experiments}
\label{app:experiments}
\subsection{Splitting}
Recall that we use $\rv{N}$ to denote a PPP, and $\rv{N}(A)$ to denote the evaluation of the PPP at some set $A \in \mathcal{F}$. The intensity function of $\rv{N}$ is given by $\lambda$, with corresponding intensity measure $\lambda(\vector{x}) \mu(d \vector{x})$. Evaluations of the intensity measure is given by
\begin{align*}
    \Lambda = \int_A \lambda(\vector{x}) \mu(d\vector{x}).
\end{align*}
Suppose we remove points from a realisation of a PPP, independently at random with probability $0 < p < 1$. The retained points are a realisation of another process $\rv{N}_p$, and the removed points are a realisation of another process $\rv{N}_{1-p}$. The intensity measures and intensity functions of each process are respectively
\begin{align*}
    \Lambda_p &= \int_A  \lambda_p(\vector{x}) \mu(d\vector{x})\\
    \Lambda_{1-p} &= \int_A \lambda_{1-p}(\vector{x}) \mu(d\vector{x}),
\end{align*}
where
\begin{align*}
    \lambda_p(\vector{x}) &= p \lambda(\vector{x})  \\
    \lambda_{1-p}(\vector{x}) &= (1-p) \lambda(\vector{x}).
\end{align*}
Moreover, $\rv{N}_p$ and $\rv{N}_{1-p}$ are independent.

In our experiments, we always artifically split realisations into training and testing data with $p=0.8$, fit an intensity function $\lambda_p$ and then evaluate the test likelihood on intensity function $\lambda_{1-p}$. 
Since we fit an intensity function $\lambda_p$ to the thinned data, then the intensity function for which we form our test evaluations is $\lambda_{1-p}(\cdot) = \frac{1-p}{p} \lambda_p(\cdot)$.

\subsection{Model settings}
All experiments are conducted on a Dual Xeon 14-core E5-2690 with 30GB of reserved RAM and a single NVidia Tesla P100 GPU.
For real data, we standardise inputs to lie in the range $[0,1]$ for all models.
For Naive, RFM and Nystr\"om models, we use the same settings as~\citet{kim2022fast}, but with an identity covariance mapping.

For SNEPPP, we use cosine activations and a Gaussian mixture model base measure. We train for 250 epochs for synthetic data and 600 epoch for real data. We use AdamW with default hyperparameters except for learning rate $0.0125/|\mathbb{X}|$, where $\mathbb{X}$ is the area of the observation window.
We use $n=200$, $n=80$ neurons and $2, 4$ mixture components respectively for real and synthetic data. 
We set $m=n$.
For the log-mixture model, we use 80 mixture components.

\subsection{Synthetic data}
We use one of the the same datasets as~\citet{kim2022fast}, in turn following~\citet{baddeley2012nonparametric}, namely data sampled from a PPP with intensity function $\lambda_{\text{GT}}(\mathbf{x}) = 0.5 \exp\big(5 - 3 \Vert \mathbf{x} - \mathcal{R}\Vert_2 \big)$, where $\mathcal{R}$ is a set of lines arranged in the shape of the letter ``R'' and $\Vert \mathbf{x} - \mathcal{R}\Vert_2$ is the shortest distance between $\mathbf{x}$ and $\mathcal{R}$. 
The dataset is sampled $100$ times and average performance is measured across the different datasets.
Each dataset has a n average size of about $N=589$ points.
Results are given in Table~\ref{tab:synthetic}.

\begin{table*}[ht]
\centering
\fontsize{9}{10.8}
\selectfont
\begin{tabular}{c|c|c|c|c}
     & NLL (exact) & NLL (MC) & RMSE & Time (seconds) \\ \hline
     Log-linear & $3.87 \pm 0.52$ & $3.87 \pm 0.52$ & $26.66 \pm 2.41$ & $3.20 \pm 0.16$ \\
     Log-linear mixture & $2.94 \pm 0.06$ & $3.92 \pm 0.06$ & $12.20 \pm 0.23$ & $3.24 \pm 0.11$ \\
     SNEPPP & $\mathbf{2.91 \pm 0.07^\ast}$ & $\mathbf{2.87 \pm 0.07}$ & $\mathbf{10.56 \pm 0.74^\ast}$ & $3.09 \pm 0.09$ \\
     RFM & n/a & $2.90 \pm 0.07$ & $11.36 \pm 0.73$ & $4.32 \pm 0.42$ \\
     Nystr\"om & n/a & $\mathbf{2.86 \pm 0.06}$ & $10.78 \pm 0.45$ & $4.19 \pm 0.24$ \\
     Naive & n/a & $2.88 \pm 0.07$ & $10.79 \pm 0.41$ & $4.28 \pm 0.30$
\end{tabular}
\caption{Synthetic experiment results. Shown are means $\pm$ sample standard deviations over $100$ randomly generated synthetic datasets and parameter initialisations. Asterisked and bold values are significantly better than closest competitor according to a two sample t-test. Bold numbers indicate values which are not significantly different to the best. Some methods do not provide exact integrated intensity functions or NLL, so we resort to Monte Carlo (MC) estimation. \label{tab:synthetic}}
\end{table*}

\subsection{Real data}
Following~\citet{kim2022fast}, we use three datasets available in the R package spatstat~\citep{baddeley2005spatstat}.
We discard covariate information in these datasets.
Bei shows the locations of $N=3605$ trees belonging to the species \emph{Beilschmiedia pendula}~\citep{sp1983diversity}. Clmfires consists of the locations of forest fires in a region of Spain, which following the benchmark we restricted to events in a rectangular region, resulting in $N=4241$ points~\citep{kim2022fast}.
Finally copper contains $N=67$ locations of copper ore deposits~\citep{berman1986testing}.

\begin{table*}
\centering
\fontsize{9}{10.8}
\selectfont
\begin{tabular}{c|c|c|c|c}
                   & NLL (exact)              & NLL (MC)                 & Count percent error      & Time (seconds)  \\ \hline
Log-linear         & $\mathbf{0.21 \pm 0.06}$ & $0.77 \pm 0.17$          & $\mathbf{0.26 \pm 0.16}$ & $2.38 \pm 0.13$ \\
Log-linear mixture & $\mathbf{0.19 \pm 0.02}$ & $0.71 \pm 0.17$          & $\mathbf{0.26 \pm 0.16}$ & $2.38 \pm 0.13$ \\
SNEPPP             & $0.85 \pm 0.90$          & $0.81 \pm 0.90$          & $\mathbf{0.29 \pm 0.25}$ & $5.89 \pm 0.44$ \\
RFM                & $-$                      & $\mathbf{0.19 \pm 0.12}$ & $\mathbf{0.19 \pm 0.13}$ & $4.34 \pm 0.30$ \\
Nystr\"om            & $-$                      & $\mathbf{0.29 \pm 0.24}$ & $\mathbf{0.30 \pm 0.17}$ & $3.96 \pm 0.27$ \\
Naive              & $-$                      & $\mathbf{0.31 \pm 0.31}$ & $\mathbf{0.26 \pm 0.17}$ & $4.30 \pm 0.30$
\end{tabular}
\caption{Copper real data experiment results. This dataset is very small ($N=67$), so we expect that simple models will outperform deep learning models, which is indeed the case. In this (and only this) dataset, the exact and MC NLLs for the log-linear and log-linear mixture models strongly disagree, perhaps because the dataset is small and the exponential function induces a high variance in the MC estimate.
\label{tab:copper} }
\end{table*}

\subsection{Case study on wildfire data}
\paragraph{Spatial fit}
We train the finite feature SNEPPP model for 20000 epochs, and train the Naive method using the same settings of~\citep{kim2022fast} but with 1000 epochs instead of 500 epochs for hyperparameter selection. 
We are more interested in the time complexity, shape and scaling of the left hand side of Figure~\ref{fig:spatial_time} than the exact constants involved in the time complexity. 
We expect that Naive methods should display a cubic time complexity in the number of datapoints, whereas finite feature methods should display a linear time complexity. This is indeed the case.

\paragraph{Spatio-temporal fit}
In order to encourage the intensity function to vary with time, we processed all dates and times to be a real number in the interval $[0, 100]$, where $0$ is the first recorded event and $100$ is the last recorded event.
We trained our models using a partial observation window in time, over the region $[0,50] \bigcup [55,100]$.
We additionally apply $p$-thinning to obtain training and testing sets, with a value of $p=0.999$ (this results in roughly 100,000 testing examples).
We use $m=n=300$ with a product-space domain and ReLU activations for each of the spatial and temporal domains.
We compare two SNEPPs --- one with ReLU and the other with exponential activations for the spatial domain --- with a baseline log-linear model ($n=m=1$ and $\sigma=\exp)$ and a log-linear mixture model ($n=m=300$, $\sigma=\exp$ and $\matrix{M}$ constrained to be diagonal with nonnegative entries). 
All activations on the temporal domain are the ReLU.
All models are trained using AdamW with default hyperparameters for 8 hours, which allows for 52 epochs over the full training set.  
For models which use $\sigma=\exp$, we found it necessary to initialise $\matrix{W}$ with a very small variance to avoid training instability.
We therefore initialise $\matrix{W}$ with a standard deviation of $10^{=4}$ for models with $\sigma = \exp$, and otherwise initialise $\matrix{W}$ from a standard Gaussian distribution. 

SNEPPP with ReLU activations achieves an NLL of $5.42$, SNEPPP with $\exp$ activations achieves an NLL of $5.85$, SNEPPP with $\exp$ activations and diagonal $\matrix{M}$ (log-linear mixture) achieves a NLL of $6.61$, and SNEPPP with $\exp$ activation and a single neuron (log-linear) achieves a NLL of $99.16$. 
Training the log-linear and log-linear mixture models is unstable due to the exponential function, so in these two models we take the lowest test NLL over training for fair comparison. 
A video showing the fits of the ReLU SNEPPP model varying in time and space are provided in the supplementary files.

\end{document}